\documentclass[runningheads]{llncs}
\usepackage[T1]{fontenc}
\usepackage{graphicx}
\usepackage{amsmath}
\usepackage{amssymb}
\usepackage{ulem}
 
\usepackage{booktabs}
\usepackage{tabularx}
\usepackage{multirow}
\usepackage[table,xcdraw]{xcolor}
\usepackage{caption}
\usepackage{cite}
\usepackage{authblk}
\usepackage{footmisc}

\begin{document}
\title{Cross-Dimensional Medical Self-Supervised Representation Learning Based on a Pseudo-3D Transformation}
\titlerunning{Cross-Dimensional SSL based on P3D Transformation}
%
\renewcommand{\thefootnote}{\fnsymbol{footnote}}

\author{Fei Gao\inst{1}$^*$ \and
Siwen Wang\inst{2}$^*$ \and
Fandong Zhang\inst{2} \and
Hong-Yu Zhou\inst{3} \and
Yizhou Wang\inst{4,5} \and
Churan Wang\inst{1}$^\dagger$ \and
Gang Yu\inst{6}$^\dagger$ \and
Yizhou Yu\inst{7}$^\dagger$
} 
%
\authorrunning{Fei Gao et al.}
%
\institute{School of Computer Science, Peking University, Beijing, China \and
Deepwise AI Lab, Beijing, China \and
Department of Biomedical Informatics, Harvard Medical School, Boston, USA \and
CFCS, School of Computer Science, Peking University, Beijing, China \and
Institute for Artificial Intelligence, Peking University, Beijing, China \and
Children's Hospital of Zhejiang University School of Medicine, Hangzhou, China \and
Department of Computer Science, The University of Hong Kong, Hong Kong
}

\maketitle              

\def\thefootnote{$*$}\footnotetext{Equal contributions.}
\def\thefootnote{$\dagger$}\footnotetext{Corresponding authors.}

\begin{abstract}
Medical image analysis suffers from a shortage of data, whether annotated or not. This becomes even more pronounced when it comes to 3D medical images. Self-Supervised Learning (SSL) can partially ease this situation by using unlabeled data. However, most existing SSL methods can only make use of data in a single dimensionality (e.g. 2D or 3D), and are incapable of enlarging the training dataset by using data with differing dimensionalities jointly. In this paper, we propose a new cross-dimensional SSL framework based on a pseudo-3D transformation (CDSSL-P3D), that can leverage both 2D and 3D data for joint pre-training. Specifically, we introduce an image transformation based on the im2col algorithm, which converts 2D images into a format consistent with 3D data. This transformation enables seamless integration of 2D and 3D data, and facilitates cross-dimensional self-supervised learning for 3D medical image analysis. We run extensive experiments on 13 downstream tasks, including 2D and 3D classification and segmentation. The results indicate that our CDSSL-P3D achieves superior performance, outperforming other advanced SSL methods.

\end{abstract}

\section{Introduction}
Medical image analysis often suffers from the lack of high-quality annotated data, which hinders the development of this field. This is primarily due to the labor-intensive and time-consuming nature of data annotation, especially for 3D data such as CT and MRI scans. Recently, self-supervised learning (SSL) has emerged as a promising approach to reduce the demand for annotated data by leveraging unlabeled data for representation learning\cite{chaitanya2020contrastive, taleb20203d, ye2022desd, haghighi2021transferable, zhou2023pcrlv2, goncharov2023vox2vec}.

However, most existing self-supervised methods are typically confined to training on either 2D or 3D data exclusively, due to the dimensionality disparity. The integration of 2D and 3D data for joint self-supervised training could substantially increase the amount of pre-training data, potentially enhancing the quality of representation learning for 3D medical image analysis. There have been a few efforts to tackle this challenge. UniMiSS\cite{xie2022unimiss} proposes the adoption of a switchable patch embedding module to accommodate both 2D and 3D inputs. Nevertheless, this approach is only applicable to transformers and not compatible with CNN models. Note that CNN is also a powerful neural architecture, widely used in a variety of medical image analysis applications. Nguyen et al.\cite{nguyen2023joint} proposed to integrate 2D CNNs with deformable attention transformers, which can simultaneously extract 2D and 3D features. However, this approach results in a disjoint representation of 2D and 3D features, and the overall rigid architecture is not compatible with existing optimized neural architectures.

To tackle the aforementioned issues, we propose a \textbf{C}ross-\textbf{D}imensional \textbf{S}elf-\textbf{S}upervised \textbf{L}earning framework based on a \textbf{P}seudo-\textbf{3D} transformation, referred to as \textbf{CDSSL-P3D}. This framework overcomes the limitations imposed by dimensional disparity and is not confined to specific neural architectures, making it a genuinely cross-dimensional SSL strategy. Specifically, drawing inspiration from the im2col\cite{chellapilla2006high,tsai2016performance} technique employed in convolution computations, we transform 2D images by sliding a window across them and unfolding the regions within each window into columns. This approach enables us to treat 2D images as 3D data. After this transformation, both 2D and 3D images can be concurrently fed into a neural network without necessitating any modifications to the architecture itself. Consequently, this seamless integration permits the direct application of existing SSL methods for the purpose of pre-training a 3D model. In our experiments, we adopt the pretext tasks proposed by \cite{zhou2023pcrlv2}, which preserves both pixelwise and semantic information in representation. 
We conduct model pre-training on a dataset comprising 6,453 3D volumes and 377,088 X-ray images. With the inclusion of the substantial collection of X-ray images into the training set, there is a noticeable improvement in performance for downstream tasks of 3D classification and segmentation. As an additional benefit, performance on 2D classification tasks can also be improved.

Overall, our contributions in this paper can be summarized as follows: (1) We propose a novel approach (CDSSL-P3D) based on the im2col transformation to tackle the challenge imposed by joint self-supervised pre-training using both 2D and 3D data, making SSL more flexible with respect to image dimensionality. (2) Our proposed method is compatible with the full spectrum of CNN and transformer architectures, not restricted solely to a specific neural architecture. (3) Our CDSSL-P3D method achieves significant performance improvements across 13 downstream tasks, including 3D medical image classification and segmentation as well as 2D medical image classification.

\section{Method}

\begin{figure}
    \centering
    \includegraphics[height=0.5\linewidth]{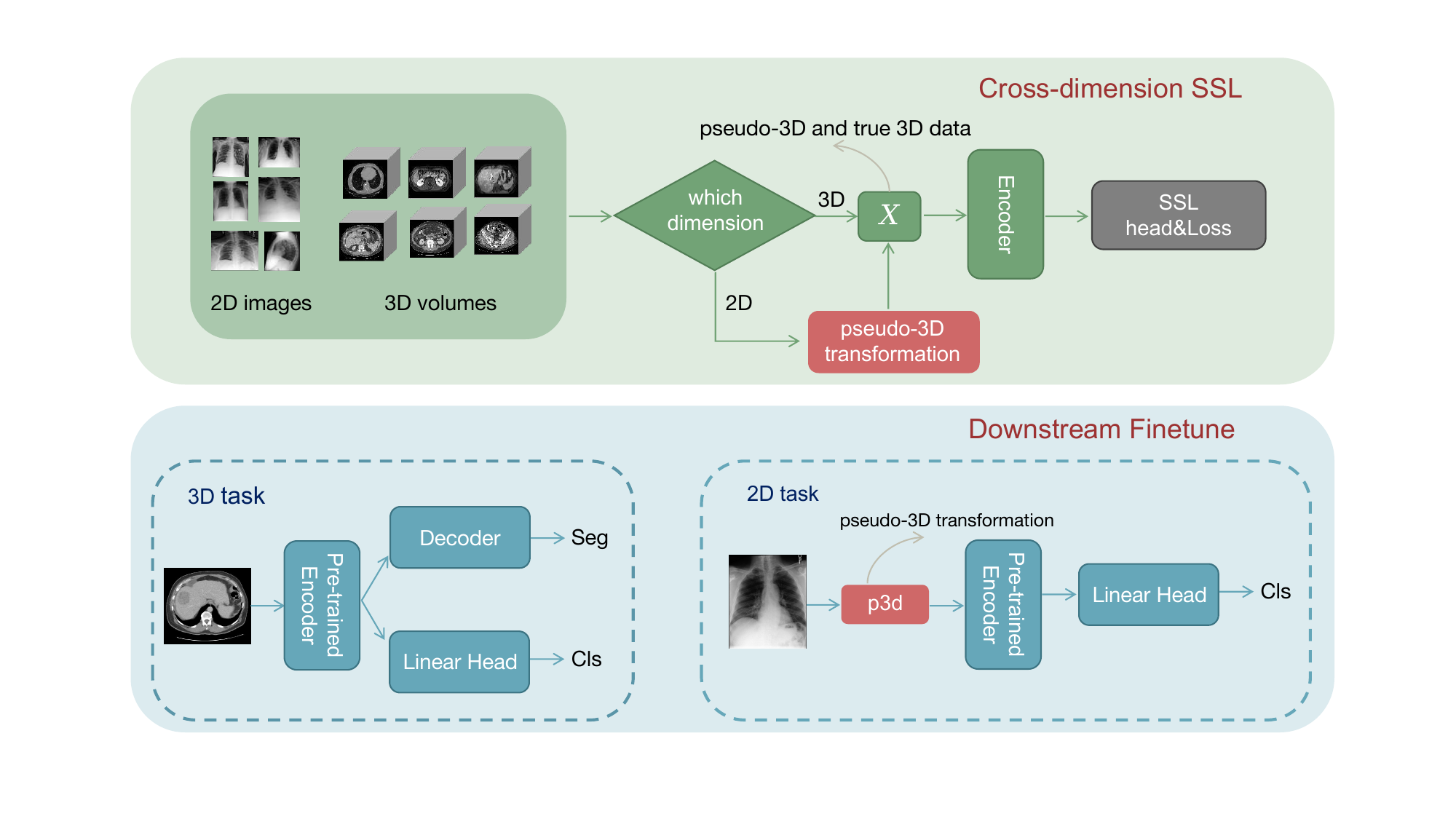}
    \caption{The overall CDSSL-P3D framework. In the pre-training stage, 2D images are converted to pseudo-3D images. Then, SSL is performed on the joint pseudo-3D and true 3D data. During the fine-tuning stage, this pre-trained 3D model is primarily used for downstream 3D tasks. As an additional benefit, downstream 2D classification tasks can be supported, and images in such 2D tasks go through our pseudo-3D transformation before fed into the 3D model.}
    \label{fig:overview}
    \vspace{-0.5cm}
\end{figure}

The overall process is shown in Fig.~\ref{fig:overview}. During the pre-training phase, we initially employ a pseudo-3D transformation to convert all 2D images into a 3D format. Subsequently, we conduct self-supervised training of a 3D model using both genuine 3D images and the transformed pseudo-3D images. This approach enables cross-dimensional representation learning. This pre-trained 3D model can be applied to 3D downstream tasks as usual. As an additional benefit, 2D classification tasks can also be carried out by converting 2D images to 3D format first using our pseudo-3D transformation. Inspired by the im2col transformation used in convolution operators, we propose a pseudo-3D transformation in a similar manner to convert 2D images into a 3D format.

\begin{figure}
    \centering
    \vspace{-0.3cm}
    \begin{center}
        \includegraphics[height=0.5\linewidth]{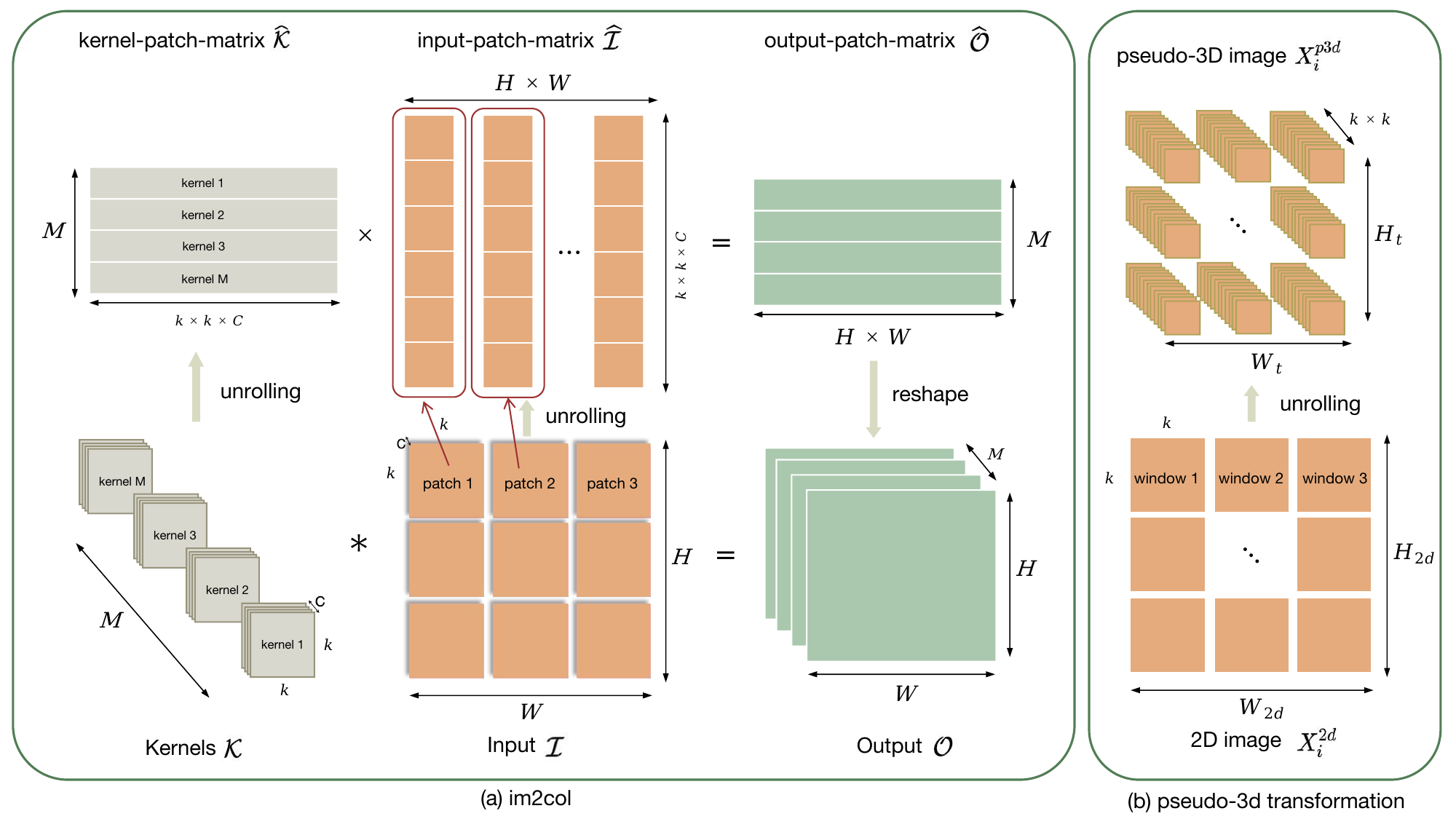}
    \end{center}
    \vspace{-0.1cm}
    \caption{The proposed pseudo-3D transformation inspired by im2col for MCMK problem. (a) Detailed depiction of im2col. Input image $\mathcal{I}$ and convolution kernel $\mathcal{K}$ are first unrolled into matrices $\widehat{\mathcal{I}}$ and $\widehat{\mathcal{K}}$, which are then multiplied to obtain the output. (b) Pseudo-3D transformation. Inspired by the transformation of $\widehat{\mathcal{I}}$, every instance of a sliding window over the entire 2D image $X^{2d}_i$ is unrolled to obtain the pseudo-3D image $X^{p3d}_i$.\vspace{-0.6cm}}
    \label{fig:im2col}
    \vspace{-0.0cm}
\end{figure}

\vspace{-3mm}
\subsection{Preliminary: image-to-column transformation (im2col)}
The application of the im2col\cite{chellapilla2006high, tsai2016performance} method has been thoroughly investigated for its effectiveness in transforming the Multiple Channel Multiple Kernel (MCMK) problem into the framework of General Matrix Multiplication (GEMM). It is widely used to accelerate convolution computation. As shown in Fig~\ref{fig:im2col}, assume an input $\mathcal{I} \in \mathbb{R}^{H\times W\times C}$ and $M$ kernels $\mathcal{K} \in \mathbb{R}^{M\times k\times k\times C}$. From the input $\mathcal{I}$ we could construct a new \textit{input-patch-matrix} $\widehat{\mathcal{I}}$ by copying \textit{patches} out of the input and unrolling them into columns of $\widehat{\mathcal{I}}$. These patches are formed in the shape of the kernel (i.e. $k\times k\times C$) at every location in the input where the kernel is to be applied. Afterwards, the dimension of the transformed input-patch-matrix $\widehat{\mathcal{I}}$ will be $H_p\times W_p$ as: 
\begin{equation}
    H_p = k\times k\times C
\end{equation}
\begin{equation}
    W_p = (\frac{H+2P-k}{s}+1) \cdot (\frac{W+2P-k}{s}+1)
\label{eq:w_im2col}
\end{equation}
where $P$ and $s$ are padding and stride in convolution.

Once the input-patch-matrix $\widehat{\mathcal{I}}$ is formed, we construct the kernel-patch-matrix $\widehat{\mathcal{K}}$ by unrolling each of the M kernels of shape $k\times k\times C$ into one row of $\widehat{\mathcal{K}}$. The shape of the resulting matrix $\widehat{\mathcal{K}}$ is $M\times (k\times k\times C)$. Then we simply perform a GEMM between $\widehat{\mathcal{K}}$ and $\widehat{\mathcal{I}}$ to obtain the output $\widehat{\mathcal{O}} \in \mathbb{R}^{H\times W\times M}$.

\subsection{Pseudo-3D transformation based on im2col}
\subsubsection{Notation:} We assume to be given 2D and 3D datasets $\{\mathcal{D}^{2d}, \mathcal{D}^{3d}\}$ where $\mathcal{D}^{2d}=\{X^{2d}_{i}\}, i \in [1,N^{2d}]$ and $\mathcal{D}^{3d}=\{X^{3d}_{i}\}, i \in [1,N^{3d}]$. Denote 2D image $X^{2d}_{i} \in \mathbb{R}^{H_{2d}\times W_{2d}}$ and 3D volume $X^{3d}_{i} \in \mathbb{R}^{H_{3d}\times W_{3d}\times D_{3d}}$. 

Intuitively, transforming input image $\mathcal{I}$ into input-patch-matrix $\mathcal{\widehat{I}}$ via im2col motivates us to conceive a pseudo-3D transformation on 2D images. Specifically, given a window size $k\times k$ and stride $s$, similar to a convolution kernel in im2col, a 2D image $X^{2d}_{i}$ can be transformed to $X^{p3d}_{i} \in \mathbb{R}^{H_{t}\times W_{t}\times D_{t}}$, where 
\begin{equation}
    H_{t}=\frac{H_{2d}-k}{s}+1, W_{t}=\frac{W_{2d}-k}{s}+1, D_{t}=k\times k.
\end{equation}
This transformation is slightly different from construction of input-patch-matrix $\mathcal{\widehat{I}}$ in im2col. In our strategy, the windows across rows and columns of $X^{2d}_{i}$ are maintained in two dimension of $H_{t}$ and $W_t$, instead of one dimension as $W_p$ (Eq.~\ref{eq:w_im2col}). Through such pseudo-3D transformation, the original information in 2D space is converted to 3D space. Such transformed data can be potentially beneficial to capture complex 3D structure and texture representations in 3D medical images for any 3D model. Using the proposed pseudo-3D transformation, we are capable of generating large-scale 3D datasets that significantly surpass the scale of existing publicly available 3D medical data. And, any 3D network can be trained concurrently on both pseudo-3D data and true 3D volume data, ensuring a seamless integration and finally a cross-dimensional SSL framework. 

\subsection{Learning Objective}
\label{Learning Objective}
The essence of this paper lies in cross-dimensional self-supervised learning, thus it is not confined to any specific self-supervised training strategy; in principle, all existing strategies are viable. We have opted for the relatively recent and powerful approach PCRLv2\cite{zhou2023pcrlv2} as an example in this study. PCRLv2 addresses information preservation in self-supervised visual representations from three aspects: pixels, semantics, and scales. First, a pixel-level objective of reconstructing the precise pixel-level details from corrupted inputs could force the model to capture pixel information in feature representations. Second, high-level siamese feature comparison is adopted to preserve semantic information in latent representations. In addition, multi-scale reconstruction and feature comparison are conducted to learn multi-scale representations.

\vspace{-4mm}
\subsection{Network}
Our approach is not constrained to any specific neural architecture and is compatible with both prevailing architectures, CNNs and Transformers. For CNN, we use a 3D version of ResNet-18~\cite{he2016deep} as the encoder, a commonly used and efficient network. In the pre-training stage, a U-like architecture with the encoder stacked with a CNN-based decoder is adopted. Regarding transformer, we adopt the pyramid vision transformer (PVT) designed for large-scale vision tasks~\cite{wang2021pyramid}, which is also adopted in \cite{xie2022unimiss} for cross-dimensional SSL. Unlike that in \cite{xie2022unimiss}, our patch embedding strategy does not necessitate switching based on data dimensions, resulting in a unified and simplified model structure. Specifically, we conduct experiments with PVT-small in this study.

\section{Experiments}

\subsection{Datasets}
\paragraph{Pre-training datasets.}
We collect 6,453 3D CT and MRI volumes from eight public datasets (LUNA16\cite{setio2017validation}, RibFrac\cite{jin2020deep}, TCIA Covid19\cite{an2020tcia}, AMOS22\cite{ji2022amos}, ISLES2022\cite{hernandez2022isles}, AbdomenCT-1K\cite{ma2021abdomenct}, Totalsegmentor\cite{wasserthal2023totalsegmentator}, Verse 2020\cite{loffler2020vertebral}) and 377,088 2D X-ray images from the MIMIC-CXR dataset\cite{johnson2019mimic} for cross-dimensional self-supervised learning.
\vspace{-3mm}
\paragraph{Downstream datasets.}
To thoroughly evaluate the effectiveness of the pre-training, we conducted comparative experiments across 13 downstream tasks. These tasks can be categorized into the following groups: (1) 3D classification (MedMNIST v2\cite{yang2023medmnist}, including six individual 3D tasks of various medical images), (2) 3D segmentation (comprising six datasets of the Liver, Hepatic Vessel (HepaV), Pancreas, Colon, Lung, and Spleen dataset from Medical Segmentation Decathlon (MSD)\cite{antonelli2022medical}), (3) 2D classification (NIH ChestX-ray)\cite{wang2017chestx}. 
\vspace{-5mm}
\begin{table}[]
\caption{Classification results of AUC on the test sets of the six 3D image datasets from MedMNIST v2~\cite{yang2023medmnist}. The results of ResNet-18+3D is taken from the original paper\cite{yang2023medmnist}. The best results of each backbone are bolded and the second-best are \uline{underlined}.}
\centering
\begin{tabular}{l|c|l|l|l|l|l|l|l}
\hline
Method       & Backbone                      & organ          & nodule          & fracture         & adrenal         & vessel          & synapse          & average        \\ \hline
ResNet-18+3D\cite{yang2023medmnist} &        & \uline{0.996}  & 0.863           & 0.712            & 0.827           & 0.874           & \uline{0.820}    & 0.848          \\
DINO\cite{caron2021emerging}  &              & 0.995          & 0.890           & 0.707            & 0.847           & 0.918           & 0.810            & 0.861          \\
SimSiam\cite{chen2021exploring} &            & 0.995          & 0.874           & 0.738            & 0.837           & 0.876           & 0.765            & 0.848          \\
TransVW\cite{haghighi2021transferable}  &    & \textbf{0.998} & \uline{0.898}   & 0.731            & 0.835           & 0.905           & 0.811            & 0.863               \\
PCRLv2\cite{zhou2023pcrlv2}       &          & 0.995          & 0.894           & \uline{0.740}    & \uline{0.853}   & \uline{0.930}   & 0.798            & \uline{0.868}          \\
CDSSL-P3D      & \multirow{-6}{*}{ResNet-18} & \textbf{0.998} & \textbf{0.908}  & \textbf{0.754}   & \textbf{0.880}  & \textbf{0.947}  & \textbf{0.835}   & \textbf{0.888}          \\ \hline
Rand. init.          &                       & 0.980          & 0.876           & 0.651            & 0.824           & 0.907           & 0.770            & 0.835               \\
UniMiSS\cite{xie2022unimiss} &               & \textbf{0.996}  & \uline{0.894}   & \uline{0.724}    & \uline{0.853}   & \uline{0.927}   & \uline{0.847}    & \uline{0.874}               \\
CDSSL-P3D      & \multirow{-3}{*}{PVT-small} & \uline{0.993} & \textbf{0.930}  & \textbf{0.761}   & \textbf{0.857}  & \textbf{0.960}  & \textbf{0.912}   & \textbf{0.902}               \\ \hline
\end{tabular}
\vspace{-0.3cm}
\label{tab:3d_cls}
\end{table}

\subsection{Experimental Details}
\paragraph{Pre-training setup.}
For the 2D MIMIC-CXR dataset, each image is resized to 224 $\times$ 224 after random crop and then transformed as a pseudo-3D patch. For 3D datasets, we randomly crop a patch from the whole CT volume with size from \{$64\times 64\times 32, 96\times 96\times 48, 112\times 112\times 56, 128\times 128\times 64$\}. The cropped patches are then resized to $64\times 64\times 32$. The input patch size is determined to strike a balance between preserving sufficient information for SSL and lower computational complexity to a manageable level. As in \cite{zhou2023pcrlv2}, for a given input image, a two-stage augmentation strategy is performed to corrupted it in global and local aspects. Global augmentation includes random flip and random affine. Local augmentation includes random noise, Gaussian blur, random swap, and random gamma. 
We employ Adam as the default optimizer and a learning rate with cosine decaying initial from 1e-3. The epochs of training is empirically set to 200, with a batch size of 96. 
\vspace{-3mm}

\paragraph{Downstream training setup.}
For all the downstream tasks, only encoder is initialized from the pre-trained models. For 3D classification tasks within MedMNIST v2, the official test set is adopted to evaluate the models performance, and the performance is measured by area under the receiver operator curve (AUC). Regarding 3D segmentation tasks, we randomly split the data of each task into training, validation and test at a ratio of 7:1:2. The Dice score is employed as evaluation metric. For 2D classification of NIH ChestX-ray, the training, validation, test sets are also randomly divided as 7:1:2, and similarly, AUC is used as the performance metric. 

\begin{table}[]
\vspace{-0.5cm}
\caption{Quantitative results on six 3D segmentation datasets. We compare the Dice (\%) on each dataset and average Dice (\%) of all datasets. The best results of each backbone are highlighted in bold and the second-best are \uline{underlined}.}
\centering
\begin{tabular}{l|c|l|l|l|l|l|l|l}
\hline
Method  & Backbone                      & Liver           & HepaV             & Pancreas        & Colon           & Lung            & Spleen          & average \\ \hline
Rand. init.     &                       & 75.2            & 60.3              & 60.1            & 30.6            & 42.2            & 92.1            & 60.1    \\
DINO\cite{caron2021emerging}    &       & 76.0            & 60.8              & 61.3            & 37.5            & 45.6            & 92.0            & 62.2    \\
SimSiam\cite{chen2021exploring} &       & 76.6            & 62.3              & 61.2            & 32.5            & 46.2            & 92.0            & 61.8    \\
TransVW\cite{haghighi2021transferable}& & 76.9            & 61.1              & \uline{61.9}    & 32.7            & 46.5            & 93.4            & 62.1    \\
DeSD\cite{ye2022desd}    &              & 76.8            & 62.2              & 61.8            & 40.2            & 52.5            & 93.9            & 64.6    \\
PCRLv2\cite{zhou2023pcrlv2}  &          & \uline{79.3}    & 62.3              & 61.5            & \uline{43.2}    & \uline{54.2}    & 95.6            & \uline{66.0}    \\
vox2vec\cite{goncharov2023vox2vec} &    & 78.5            & \uline{63.8}      & 61.8            & 32.6            & 47.2            & \uline{96.1}    & 63.3    \\
CDSSL-P3D & \multirow{-8}{*}{ResNet-18}       & \textbf{81.2}   & \textbf{65.0}     & \textbf{63.0}   & \textbf{46.2}   & \textbf{57.1}   & \textbf{96.2}   & \textbf{68.1}    \\ \hline
Rand. init. &                           & 77.9            & 63.6              & 63.5            & 39.3            & 49.3            & 93.5            & 64.5    \\
UniMiSS\cite{xie2022unimiss} &          & \uline{81.1}    & \uline{64.3}      & \uline{64.1}    & \uline{44.6}    & \uline{56.7}    & \uline{95.4}    & \uline{67.7}    \\
CDSSL-P3D & \multirow{-3}{*}{PVT-small} & \textbf{82.5} & \textbf{67.8}     & \textbf{65.5}   & \textbf{50.4}   & \textbf{60.5}   & \textbf{96.2}   & \textbf{70.5}    \\ \hline
\end{tabular}
\vspace{-0.3cm}
\label{tab:3d_seg}
\end{table}

\vspace{-1mm}
\subsection{Results}
\vspace{-0mm}
\paragraph{Comparing to other SSL Methods.}
The proposed CDSSL-P3D is compared with random initialization, and seven advanced SSL methods including DINO~\cite{caron2021emerging}, SimSiam~\cite{chen2021exploring}, TransVW~\cite{haghighi2021transferable}, DeSD~\cite{ye2022desd}, PCRLv2~\cite{zhou2023pcrlv2}, vox2vec~\cite{goncharov2023vox2vec} and UniMiSS~\cite{xie2022unimiss}. Note that the first six methods use CNNs as their encoder backbone and UniMiSS\cite{xie2022unimiss} adopts a transformer as its backbone. In addition, only UniMiSS exploits cross-dimensional data as ours while the first six methods use data with a single dimensionality only. Thus, in the comparison experiments, the first six methods are pre-trained using data with the same dimensionality as the downstream tasks. Our CDSSL-P3D and UniMiSS are pre-trained on all 2D and 3D data collected for pre-training. As detailed in Tables~\ref{tab:3d_cls},~\ref{tab:3d_seg}, the proposed CDSSL-P3D is compared with the competitors primarily on 3D medical tasks including six 3D classification tasks (Table~\ref{tab:3d_cls}) and six 3D segmentation tasks (Table~\ref{tab:3d_seg}). In addition, one 2D classification task (Table~\ref{tab:2d_cls}) is also conducted because downstream 2D classification tasks are supported by our pre-trained 3D model. The following conclusions can be drawn from the tables: 1) SSL significantly enhances model performance compared with random initialization. 2) Transformer-based models generally outperform CNN-based methods. 3) Our CDSSL-P3D framework demonstrates notable performance improvements for both CNNs and Transformers, confirming the effectiveness of our cross-dimensional strategy on the two predominant neural architectures. 4) CDSSL-P3D achieves the highest performance across all tasks, surpassing the second-best methods, PCRLv2\cite{zhou2023pcrlv2} and UniMiSS\cite{xie2022unimiss}, by 2.0\%, 2.8\% (3D classification), 2.1\%, 2.7\% (3D segmentation) and 1.0\%, 1.6\% (2D classification), respectively. Note that the performance does not deteriorate with the 3D model compared with the 2D model on the NIH ChestX-ray dataset, indicating that employing pseudo-3D transformations for 2D downstream tasks does not incur a loss in performance (first and sixth rows in Table~\ref{tab:2d_cls}). Furthermore, given the substantial size of the NIH ChestX-ray dataset (over 100,000 images), we conduct additional experimental comparisons at varying training data ratios (Table~\ref{tab:2d_cls}). The results indicate that the CDSSL-P3D framework consistently provides the most significant improvement across different ratios, with particularly notable improvements at smaller training set sizes.
\vspace{-3mm}

\begin{table}[]
\vspace{-0.5cm}
\caption{Quantitative results of different SSL strategies on NIH ChestX-ray dataset for 2D Classification, measured by AUC under different ratios of training data. The best results are bolded and the second-best are \uline{underlined}.}
\centering
\begin{tabular}{l|c|l|l|l|l}
\hline
Method   & Backbone                         & 10\%           & 30\%           & 50\%           & 100\% \\ \hline
Rand. init. & \multirow{5}{*}{ResNet-18 2D}                          & 0.695          & 0.735          & 0.774          & 0.808 \\
DINO\cite{caron2021emerging}     &          & 0.723          & 0.790          & 0.787          & 0.818       \\
SimSiam\cite{chen2021exploring}  &          & 0.728          & 0.785          & 0.789          & 0.819       \\
TransVW\cite{haghighi2021transferable}  &   & 0.715          & 0.753          & 0.788          & 0.816       \\
PCRLv2\cite{zhou2023pcrlv2}   &             & \uline{0.770}  & \uline{0.809}  & \uline{0.819}  & \uline{0.828} \\ \hline
Rand. init. & \multirow{2}{*}{ResNet-18 3D}      & 0.703          & 0.745          & 0.772          & 0.805 \\
CDSSL-P3D  &                                & \textbf{0.778} & \textbf{0.818} & \textbf{0.827} & \textbf{0.838} \\ \hline
Rand. init. & \multirow{3}{*}{PVT-small}  & 0.712          & 0.764          & 0.782 & 0.816 \\
UniMiSS\cite{xie2022unimiss}  &             & 0.771          & 0.809          & 0.820 & 0.840 \\
CDSSL-P3D  &                                & \textbf{0.789} & \textbf{0.823} & \textbf{0.840} & \textbf{0.856} \\ \hline
\end{tabular}
\vspace{-0.3cm}
\label{tab:2d_cls}
\end{table}

\vspace{-0mm}
\paragraph{Ablation of pre-training with different dimension.}
A key contribution of this paper is the joint pre-training of 2D and 3D data, which offers distinct advantages over pre-training with data from a single dimension alone. To substantiate the efficacy of joint pre-training,  we have compared the models performance pre-training with different data dimensions (2D, 3D, 2D+3D) on downstream 2D (NIH ChestX-ray) and 3D (MedMNIST v2) tasks (shorten as NIH and MedM in Table~\ref{tab:dimension},~\ref{tab:window},~\ref{tab:stride}). This comparison is conducted on ResNet-18 as a example, which is also adopted in the following ablation studies. As depicted in Table~\ref{tab:dimension}, joint pre-training exhibits a significant enhancement compared to using either 2D or 3D data exclusively.
\vspace{-3mm}

\paragraph{Ablation on window size.}
Table~\ref{tab:window} presents the results under various window sizes ($3\times 3$, $5\times 5$, $7\times 7$). Overall, larger windows tend to yield superior performance ($5\times 5$ and $7\times 7$ outperform $3\times 3$). Nevertheless, it is not the case that larger windows always lead to better results. The optimal performance is achieved with a 5x5 window size. We speculate that the advantage of larger window sizes may be attributed to the compatibility with the dimensionality of 3D data, leading to better joint pre-training integration. 
\vspace{-0mm}

\begin{minipage}[c]{0.3\textwidth}
\centering
\vspace{0.3cm}
\captionof{table}{Ablation of SSL dimension.}
\vspace{-0.3cm}
\centering
\begin{tabular}{l|l|l}
\hline
dim & MedM        & NIH \\ \hline
2D            & 0.863           & 0.828           \\
3D            & 0.875           & 0.824           \\
2D+3D         & \textbf{0.888}  & \textbf{0.838 }          \\ \hline
\end{tabular}
\label{tab:dimension}
\vspace{-0.2cm}
\end{minipage}
\hspace{0.1cm}
\begin{minipage}[c]{0.3\textwidth}
\centering
\vspace{0.3cm}
\captionof{table}{Ablation of window size.}
\vspace{-0.3cm}
\centering
\begin{tabular}{l|l}
\hline
window size        & MedM          \\ \hline
$3\times 3$         & 0.876             \\
$5\times 5$         & \textbf{0.888}    \\
$7\times 7$         & 0.881             \\ \hline
\end{tabular}
\label{tab:window}
\vspace{-0.2cm}
\end{minipage}
\hspace{0.1cm}
\begin{minipage}[c]{0.3\textwidth}
\centering
\vspace{0.3cm}
\captionof{table}{Ablation of sliding stride.}
\vspace{-0.3cm}
\centering
\begin{tabular}{l|l}
\hline
stride & MedM       \\ \hline
1      & \textbf{0.888} \\
2      & 0.885          \\
3      & 0.886          \\ \hline
\end{tabular}
\label{tab:stride}
\vspace{-0.1cm}
\end{minipage}

\paragraph{Ablation on sliding stride.}
Table~\ref{tab:stride} shows the comparison of model performance under various stride settings. We evaluate the models at the optimal window size of 5x5 to assess the impact of different strides. The results indicate that the performance differences across strides are subtle, demonstrating the robustness of our approach to changes of stride.
\vspace{-2mm}

\section{Conclusion}
We propose a cross-dimension self-supervised learning strategy (CDSSL-P3D) aiming to perform jointly pre-training of 2D and 3D data in medical images. The introduced strategy is not confined to specific network architectures, which can be applied for CNNs and Transformers. We conduct experiments with CNN and Transformer on 13 downstream tasks and compare with a series of advanced SSL methods. Extensive evaluation results amply substantiate the effectiveness of CDSSL-P3D.

\begin{credits}
\subsubsection{\discintname}
The authors have no competing interests to declare that are
relevant to the content of this article.
\end{credits}

%
%

\bibliographystyle{splncs04}
\bibliography{paper} 
\end{document}